\pdfoutput=1
\documentclass[11pt]{article}

\usepackage[]{emnlp2021}

\usepackage{times}
\usepackage{latexsym}

\usepackage[T1]{fontenc}
\usepackage[utf8]{inputenc}

\usepackage{microtype}

\usepackage{booktabs}   %% For formal tables:
\usepackage{subcaption} %% For complex figures with subfigures/subcaptions
\usepackage{mathtools}
\usepackage{xspace}

\usepackage{listings}
\usepackage{mathtools}

\usepackage{amssymb}
\usepackage{enumitem}
\usepackage{outlines}
\usepackage{tikz}
\usepackage{amsmath}

\usepackage[frozencache=true,cachedir=minted-cache]{minted}
\makeatletter
\define@key{FV}{highlightlines}{\edef\FV@HighlightLinesList{#1}}
\makeatother

\usepackage{algpseudocode}
\usepackage{algorithm}

\usepackage{blindtext}
\title{Long-Range Modeling of Source Code Files with eWASH: Extended Window Access by Syntax Hierarchy}

\author{
    Colin B. Clement\\
    Microsoft Cloud and AI\\
    \texttt{colin.clement@microsoft.com}\\\And
    Shuai Lu \\
    Microsoft Research\\
    \texttt{shuailu@microsoft.com}\\\AND
    Xiaoyu Liu\\
    Microsoft Cloud and AI\\
    \texttt{lixiaoyu@microsoft.com}\\\And
    Michele Tufano\\
    Microsoft Cloud and AI\\
    \texttt{mitufano@microsoft.com}\\\AND
    Dawn Drain\\
    Microsoft Cloud and AI\\
    \texttt{dawndrain95@gmail.com}\\\And
    Nan Duan \\
    Microsoft Research\\
    \texttt{nanduan@microsoft.com}\\\AND
    Neel Sundaresan\\
    Microsoft Cloud and AI\\
    \texttt{neels@microsoft.com}\\\And
    Alexey Svyatkovskiy\\
    Microsoft Cloud and AI\\
    \texttt{alsvyatk@microsoft.com}\\
}

\begin{document}

\maketitle

% Editing commands
\newcommand{\ie}{\textit{i.e.,}~}
\newcommand{\eg}{\textit{e.g.,}~}
\newcommand{\etc}{\textit{etc.}~}
\newcommand{\etal}{\textit{et al.}~}

\begin{abstract}

Statistical language modeling and translation with transformers have found many successful applications in program understanding and generation tasks, setting high benchmarks for tools in modern software development environments. The finite context window of these neural models means, however, that they will be unable to leverage the entire relevant context of large files and packages for any given task. While there are many efforts to extend the context window, we introduce an architecture-independent approach for leveraging the syntactic hierarchies of source code for incorporating entire file-level context into a fixed-length window. Using concrete syntax trees of each source file we extract syntactic hierarchies and integrate them into context window by selectively removing from view more specific, less relevant scopes for a given task. We evaluate this approach on code generation tasks and joint translation of natural language and source code in Python programming language, achieving a new state-of-the-art in code completion and summarization for Python in the CodeXGLUE benchmark. We also introduce new CodeXGLUE benchmarks for user-experience-motivated tasks: code completion with normalized literals, method body completion/code summarization conditioned on file-level context.

\end{abstract}

\section{Introduction}

% discuss why program understanding tasks are relevant for NLP community, importance of software

% motivate the importance of extended context for program understanding and generation tasks
Large transformer models~\cite{vaswani2017attention} and the pre-training/fine-tuning paradigm~\cite{devlin2018bert,lewis2019bart,radford2018improving} have become an essential part of state of the art natural language processing. Beyond the domain of natural language, these models and procedures have enabled rapid progress in the software engineering space, including applications in code completion~\cite{gptc,svyatkovskiy2019pythia,clement2020pymt5,raychev2014code, bruch2009learning}, natural language to code (NL2Code), code feature summarization~\cite{clement2020pymt5,moreno2013automatic,scalabrino2017automatically,wan2018improving,alon2018code2seq,moreno2014automatic}, code search~\cite{husain2019codesearchnet,feng2020codebert}, unit test generation~\cite{tufano2020generating} and even bug fixing~\cite{deepdebug_java} and detection~\cite{zhai2020cpc}.

A major difference between transformers and their antecedents like recurrent neural networks (RNN) is their strictly enforced finite context window. Whereas an RNN can iteratively consume as many tokens as is required, transformers can only consume up to a finite amount decided at training time. Further, it is impractical to simply expand the window as the memory and compute requirements of the attention mechanism scale quadratically with context length. There have been efforts to economically expand the window by modifying the attention mechanism with low-rank queries and keys~\cite{beltagy2020longformer}, sparse connections~\cite{parmar2018image,zaheer2020big}, and more recently approximation by kernel methods~\cite{performer}. There have also been methods developed to condition generation on retrieved documents~\cite{lewis2020retrieval,lewis2020pre} for knowledge-intensive applications. Complimentary to these approaches, and consistent with any sequence model or architecture, we propose a method for extracting the most important features distant from the task at hand, implemented in this case using the syntax tree of source code.

A source code document has nested scopes and references to other documents/libraries, and software engineering tasks must leverage knowledge at all of these scales. Even single source code files often exceed the length of the context window, and so it is clear that progress in modeling source code requires overcoming this limitation. Instead of proposing a novel transformer architecture capable of ingesting more context, we propose a method of compressing the context of source code files using the syntax of the programming language itself. In this paper we introduce eWASH: Extended Window Access by Syntax Hierarchy, which leverages the syntax hierarchy of source code to give our models longer ranged vision by prioritizing higher level scopes and names for source code elements in a file which are not immediately in focus. Using eWASH assumes that higher level scopes like function signatures summarize the whole method, and viewed in this way eWASH could be applied to natural language by extracting key terms in a long document or summarizing features distant from the task at hand.

We start by explaining eWASH with a motivating example and define its features for three important software engineering tasks. The first is code completion, in which some number of tokens are predicted to extend a partially complete source code file. The second is method completion, wherein a whole method body is predicted from a method signature and docstring. The third is code summarization or docstring generation, wherein a method is mapped to a natural language docstring. We then discuss the Python training data and the models employed in this study, which include the auto-regressive GPT-C~\cite{gptc}, an eWASH extended version called XGPT-C, Python method/docstring prediction model PyMT5~\cite{clement2020pymt5}, similarly named XPyMT5, and Performer~\cite{performer} and Reformer~\cite{reformer} baselines. We demonstrate through experiments the value of eWASH for the source code domain with state of the art performance for code completion and method completion and code summarization or docstring generation. Finally, we study user-experience motivated properties of XGPT-C, and extend the source code benchmark set CodeXGLUE~\cite{codexglue} with two new tasks: literal-normalized code completion and method completion. Surprisingly, eWASH excels at code completion even when the context window is not a limiting factor.

\iffalse
% name of the approach? 
We evaluate the proposed approach on the code completion, code summarization and natural language to code tasks. Code completion has made substantial progress in recent years. From single-token completion of member-access expressions and method invocations via sequence classification [cite] and re-ranking [cite] to correctly filling in arguments [cite], or entire line statements [cite] or snippets [cite]. Nowadays, software developers are coding in multiple languages, which brings up a need for multi-lingual code completion systems. The simple approach of deploying a separate mono-lingual model for each programming language is not scalable and resource hungry.  
% Discuss what GPT-C does not do with respect to extended context
Code summarization and joint modeling of natural language and source code has been extensively studied in [cite]. These works, however, disregard the class level context when generating method bodies based on signatures and docstrings. 
% Discuss what PyMT5 does not do with respect to extended context
\fi

\begin{figure*}[htb]
    \centering
    \includegraphics[width=0.95\textwidth]{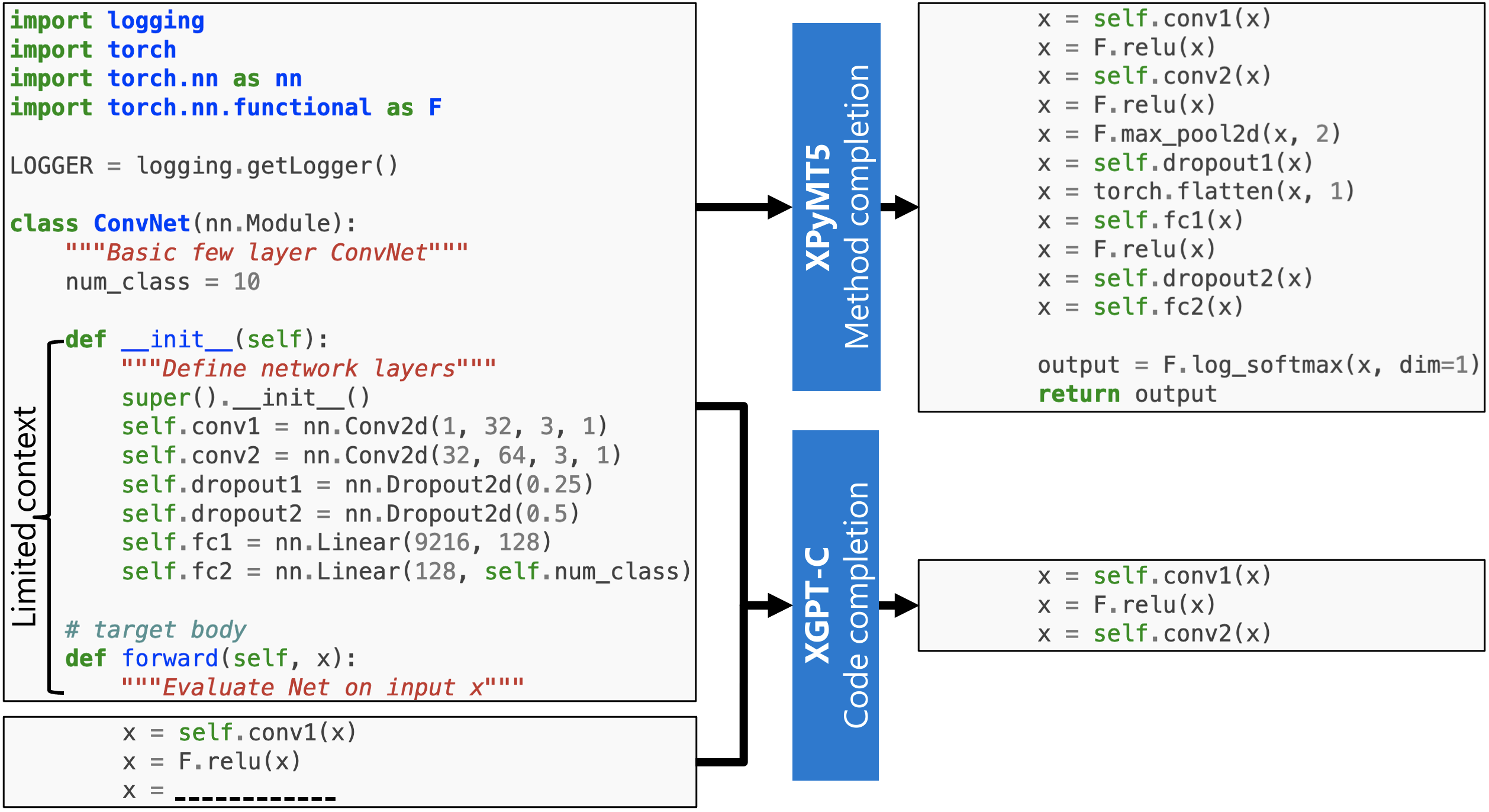}
    \caption{
    An example scenario where both method completion (top right) and code completion (bottom right) are performed by our eWASH models XPyMT5 and XGPT-C, respectively. Method completion aims to predict a whole method body conditioned on a signature and docstring and other context, and code completion aims to predict any number of tokens to complete a member, line, or even a scope context conditioned on any incomplete code string. In this case XPyMT5 is tasked with predicting the whole body of \mintinline{python}{ConvNet.forward}. In this case it's clear that both the layers assigned in \mintinline{python}{ConvNet.__init__} and the import statements above are important information. In another case XGPT-C aims to complete the assignment, and again the class attributes and import statements are important. Both models have a limit context window illustrated at left, and exclude important information.
    }
    \label{fig:ec-example}
\end{figure*}

\section{Motivating example}

In this paper we consider three software engineering tasks: code completion, method completion, and docstring completion or code summarization. Code completion is the auto-regressive prediction of one or more tokens conditioned on a provided context. Figure~\ref{fig:ec-example} shows an incomplete Python program implementing a neural network. A developer would like a model which can predict the body of \mintinline{python}{ConvNet.forward} (method completion) by composing the layers defined in \mintinline{python}{ConvNet.__init__} and globally imported operations. There is a limited context window, illustrated at the left of the figure, and so while the model can (in this fortunate case) see the layer definitions, it is ignorant of the imports and the global \mintinline{py}{LOGGER} object.

In many cases predicting a whole method is not easy, but asking for a few members or a completed line~\cite{gptc} is still desirable. The bottom of fig.~\ref{fig:ec-example} shows the code completion task finishing an assignment and completing a line in a partial implementation of  \mintinline{python}{ConvNet.forward}. Here, again, crucial information is missing from the input of the model, which will prevent it from being able to reference the imports. Further, one can easily imagine a scenario in which more spurious information is fed into the model instead of, for example, the definitions of the neural network layers in the  \mintinline{py}{__init__}. How can we ensure the model is shown important information for predictions?

\subsection{Extended Window Access by Syntax Hierarchy}

Software developers carefully organize their code into scopes and portable elements using hierarchies like methods and classes, so we hypothesize that labels of these scopes are more important for long-range modeling than their contents. We propose eWASH, Extended Window Access by Syntax Hierarchy, in which we compress the context provided to our model by prioritizing, for example, function signatures over function bodies. Since most code is written inside methods, we center method bodies as the focus of the modeling task for eWASH, calling each method being modeled the `focal method.'

\begin{figure}[htb]
    \centering
    \includegraphics[width=0.95\columnwidth]{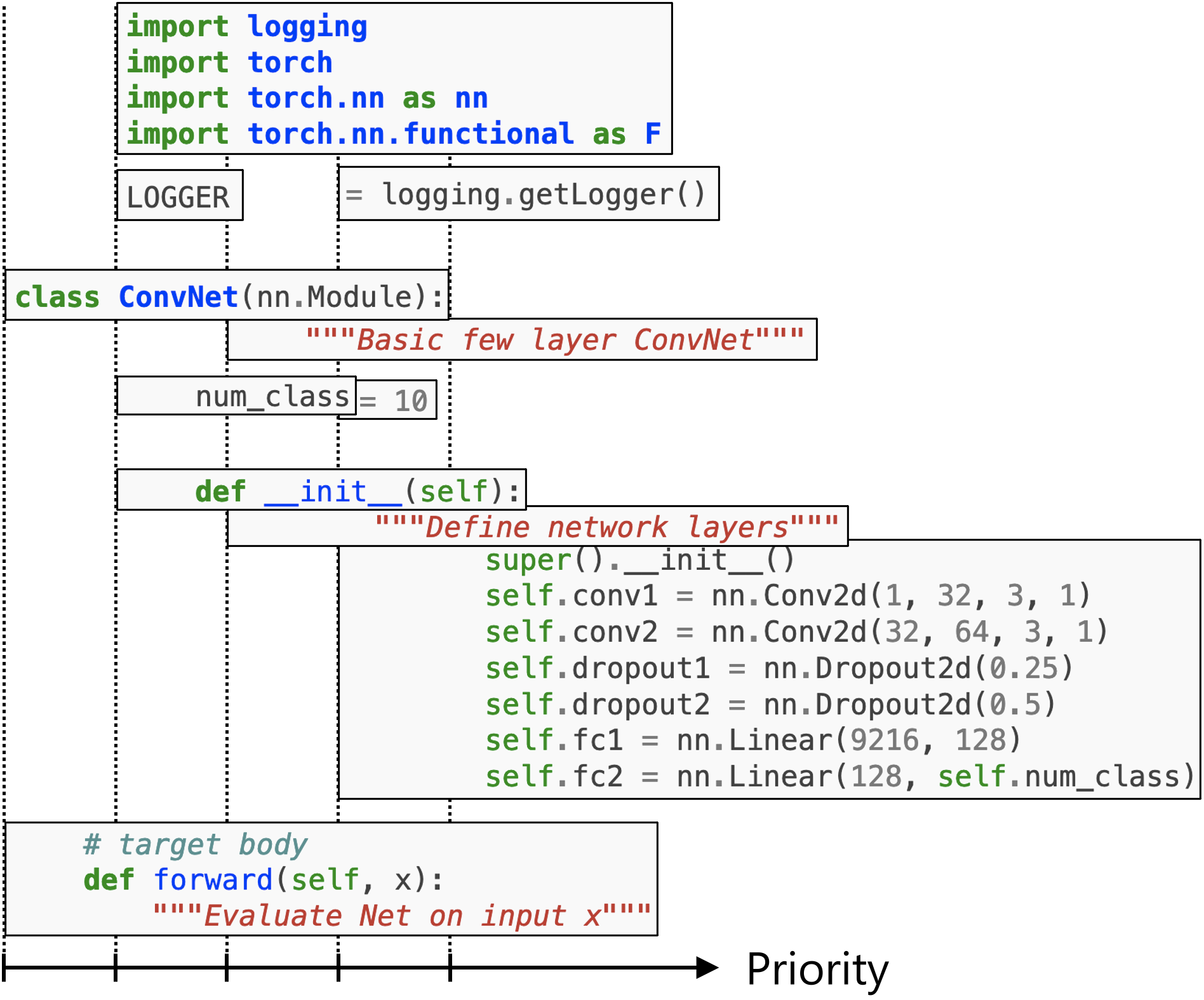}
    \caption{
    An illustration of the syntax hierarchy of the input context in the method completion example from Fig.~\ref{fig:ec-example}. The eWASH (Extended Window Access by Syntax Hierarchy) method selectively fills the model context going down the file, in the order of priority level indicated, and stops when the token budget of the model context is filled. eWASH presupposes that names of entities at higher scopes are more relevant to the task at hand than entities at lower scopes.
    }
    \label{fig:syntax-hierarchy}
\end{figure}

Figure~\ref{fig:syntax-hierarchy} shows how eWASH uses syntax hierarchies to prioritize elements of the context of the method and code completion example of Fig.~\ref{fig:ec-example}. The focal method in this case is \mintinline{py}{ConvNet.forward}, but could be any other method in the module. The most important part for modeling the body of this focal method is its signature and docstring (if present) and containing class definition (if the focal method is a class method). After this we prioritize global import statements and assigned values (but not yet the assigned expressions), followed by class attributes, peer class method signatures, class docstring, peer class method docstrings, and finally global expressions and the code bodies of peer class methods.

In practice, eWASH is implemented by taking the concrete syntax tree of the source file and organizing the syntactic elements in our priority list, tokenizing each element, and then descending the priority list, taking elements until the context window has been filled. For training the method completion of XPyMT5, we arrange the eWASH context in the input with a control code to indicate which method is to be completed (\texttt{\# target body} in Fig.~\ref{fig:ec-example}), and we arrange the target to be the method body. eWASH yields $N$ total training samples from a file with $N$ total methods and class methods. For docstring completion or code summarization, the source contains the method signature and body, and the target contains the desired docstring, and a control code is used to instruct the model which task it is to perform, just like PyMT5~\cite{clement2020pymt5}.

For code completion, as we use an auto-regressive decoder in the form of XGPT-C there is no special `position,' and so we create a rolling window across the focal method body. We reserve 3/4 (768/1024 tokens) of the tokens for the context, and 1/4 (256/1024 tokens) for the rolling window of the body. In the case of a method which exceeds 256 tokens, the training sample for that method is decomposed into multiple `windows,' and one file yields at least $N$ training samples for a file with $N$ method and class method definitions.

\begin{table*}[htb]
\centering
\begin{tabular}{llllllllll} \toprule
	\textbf{Model}   &  \textbf{PPL} & \bf ROUGE-L Prec. & \bf Recall & \textbf{Edit dist.} & \textbf{EM@5 (\%)}  & \textbf{Size} \\\midrule
%			\multicolumn{2}{c}{\textbf{ROUGE-L}} & \textbf{Edit dist.} & \textbf{EM@5 (\%)}  & \textbf{Size} \\ \cmidrule{3-4} \cmidrule{8-10}
%	                & & Precision &  Recall \\ \midrule
Performer & 2.06 & 0.69 & 0.80 & 85.1 & 41.2 & 125M& \\
Reformer & 2.02 & 0.70 & 0.81 & 86.3 & 46.1 & 116M & \\
XGPT-C & \bf 1.35 & \bf 0.85 & \bf 0.93 & \bf 90.8 & \bf 49.4 & 125M\\\midrule
GPT-C, Norm Literals & 1.83 & 0.81 & 0.94 & 89.0 & 46.3 & 125M \\	%final  
XGPT-C, Norm Literals  & \bf 1.25 & \bf 0.90 & \bf 0.96 & \bf 93.7 & \bf62.4 & 125M\\	
\bottomrule
\end{tabular}
\caption{Evaluation results comparing XGPT-C decoder-only model trained with extended hierarchical context with baselines on code completion from sampled methods. Model performance metrics are reported on test samples from the CodeXGLUE code completion task as described in Sec.~\ref{sec:experimental-conditions}. 
}
\label{tab:eval_codecompletion}	
\end{table*}

\section {Dataset}\label{sec:dataset}

\subsection{Pre-training}
The data for training is the same for both XGPT-C and XPyMT5, and consists of all 5+ star GitHub repositories which are primarily Python, filtered by files which were either Python 3 compliant or were successfully fixed by \mintinline{python}{lib2to3}. Further, there was a time limit of 10 seconds placed on the parsing process to eliminate files which are essentially data files as they tend to contain very large lists for example. Table~\ref{tab:data-stats} shows summary statistics of this dataset for a sense of scale.

\subsection{Fine-Tuning and Evaluation}
For evaluation and fine-tuning of code completion we used the Py150~\cite{raychev2016probabilistic} from CodeXGLUE~\cite{codexglue}, and for method and docstring completion we used the CodeSearchNet~\cite{husain2019codesearchnet} dataset. Py150 is larger than CodeSearchNet for Python, but has selected repositories with good docstring coverage, allowing better evaluation of the method/docstring completion task.

\begin{table}[htb]
    \centering
    \small
    \begin{tabular}{l l l l l l}
        Lang & Repos & Files & Methods & Classes & Lines \\\midrule
        Python   & 238k         & 4.3M  & 15M     & 10.6M   & 1.4B
    \end{tabular}
    \caption{
    Summary statistics of our Python~parallel corpus compared to others presented in the literature. CSN contains 500k Python~ methods with docstrings, among 6 other languages. Our parallel corpus is 3$\times$ as large as the next largest, and over 15$\times$ the size of the next largest Python~parallel corpus.
    }
    \label{tab:data-stats}
\end{table}

\section{Baseline Models}

We consider state-of-the-art transformer models for code completion and code summarization tasks in the CodeXGLUE benchmark as our baselines. Namely, the generative pre-trained transformer model for code~\cite{gptc} (GPT-C) and the Python method text-to-text transfer transformer model~\cite{clement2020pymt5} (PyMT5). We also experiment with two memory-efficient transformers---Reformer~\cite{reformer} and Performer~\cite{performer}---which enables modeling of context lengths in excess of 1024 tokens.

\subsection{GPT-C}
GPT-C is an auto-regressive language model pre-trained on a large unsupervised source code corpora. Treating the source code data as a sequence of lexically-defined tokens, GPT-C extracts training samples as a sliding window from each source code file. This baseline uses an approach based on statistical language modeling of source code, with several normalization rules extracted from concrete syntax tree of a program. To overcome the issue of different styles and white space or tab conventions, it transforms the code into symbolic program tokens using custom tokenizer and regenerates the code with a common style. During pre-processing, GPT-C parses program code in each file, extracts information about token types, normalizes uncommon literals, trains a sub-token vocabulary, and encodes the files into sub-token sequences. This is done both for training and inference. GPT-C decoder-only model has about 125M parameters and a context length of 1024 tokens. % we refer to 12L model

\subsection{PyMT5}
PyMT5 is a transformer encoder-decoder model jointly pre-trained on a large-scale corpus of Python source code and natural language contained in the docstring summaries. PyMT5 training samples are supervised pairs of function code features---function signatures, docstrings and bodies---extracted by means of a parser. PyMT5 is finetuned to translate between all non-degenerate combinations of code features in a multi-modal setting, e.g. simultaneously signature and docstring to body, signature to docstring and body, signature to body, etc. PyMT5 only uses information from a single method and naturally is missing imports, peer class and method definitions, and global assignments. PyMT5 has 406M parameters and a context width of 1024 tokens for both the encoder and decoder.

\subsection{Memory-Efficient Transformers}
Reformer and Performer transformer models attempt to break the infamous quadratic attention bottleneck and allow for efficient modeling with much longer than the standard 1024 token context window. Reformer~\cite{reformer} includes three memory optimizations: reversible layers (to trade off memory with time), axial positional embeddings, and a bucketed attention. %During the backward gradient accumulation in reversible layers memory can be saved by recomputing the intermediate hidden states. Axial positional embeddings reduce the number of positional embedding parameters, and bucketed attention enforces queries to attends to the closest key-value pairs as determined by LSH buckets.
Performer~\cite{performer} develops a linear approximation to the attention layer $AV \approx \sigma(Q^TK)V$, where $K$, $Q$, and $V$ are the key, query and value matrices of the attention mechanism and $A$ is the softmax kernel approximation, and exploits the linearity to improve computational efficiency.

\section{Defining the Tasks at Hand}

\subsection{Code completion}

Code completion is the auto-regressive completion of source code tokens, as illustrated in the bottom of Fig.~\ref{fig:ec-example}. We perform code completion as defined in CodeXGLUE~\cite{codexglue} as well as using normalized literals. The literal normalization improves user experience of the code completion tool~\cite{gptc} by abstracting personally identifiable information and encouraging the model to focus on code modeling over arbitrary strings. Names, the phone number, IP addresses, and more may be preserved in the string or numeric literals. We normalize the literals in source codes to some special tokens. Considering that frequently used literals may contain useful information, e.g. \texttt{"\_\_main\_\_"} or \texttt{"utf-8"}, we preserve the 200 most frequent string and 30 most frequent numeric literals.

\subsection{Method completion}

Method completion is the prediction of a method body implementation conditioned on a signature, optional docstring, and any more context. The authors of PyMT5 performed this task using no context beyond the focal method, and XPyMT5 uses eWASH compressed file-level context. We contribute method and docstring completion conditioned on file-level information as a task to CodeXGLUE, based on the CodeSearchNet dataset in order to bolster its user-experience motivated tasks.

\subsection{Docstring Completion/Code Summarization}

Docstring completion is the prediction of a docstring conditioned on a focal method and optional context, and was also performed by PyMT5 on focal-methods alone. Code summarization is closely related, as docstrings often express a summary of their method, but also include annotated arguments, return values, exceptions, and even test cases via \texttt{doctest}. We train on docstring completion but evaluate on the CodeSearchNet dataset which attempts to remove everything but the summary which is assumed to be in the first paragraph of the docstring.

\iffalse
\subsection{Natural Language to Code}

This task aims to predict a code implementation of a natural language description of a program, and is often modeled as the reverse of docstring completion or code summarization. We expand the definition of the problem in this case to also condition on file-level eWASH context, in other words allow the model to see the file in which the method being described by natural language will be placed. We also contribute natural language to code conditioned on file-level context to CodeXGLUE, using the CodeSearchNet dataset.
\fi

\section{Model Training}

\subsection{XGPT-C}
We trained XGPT-C on the Python dataset described in \ref{sec:dataset}. Each training sample is a method body along with its corresponding extended context. In XGPT-C, we follow GPT-C~\cite{gptc}, using a multi-layer Transformer decoder as the model architecture and the causal language model as the training objective. We use 12 layers Transformer decoder, 12 attention heads with 768 hidden dimension in total and sentencepiece~\footnote{https://github.com/google/sentencepiece} BPE vocabulary of size 50,000. The total number of model parameters is 125M. The pre-training period takes 2 weeks on sixteen 32GB Tesla V100 GPUs, and all hyperparameters were left as \citet{gptc}.

\subsection{XPyMT5}
We trained XPyMT5 on the same Python dataset as XGPT-C. Similar to PyMT5, Each Python file yielded between $N$ and $3N$ training samples where $N$ is the number of methods and class methods in the file. Each method teaches the model to complete the method body conditioned on its signature (and docstring if it exists), to predict the docstring (if it exists) from the method, and to predict the whole method from just the docstring (if it exists). In this way, XPyMT5 also can jointly predict code and natural language, but we did not include all degenerate combinations like PyMT5 as the training set was already much larger due to the extended context.

XPyMT5 uses the same whitespace-augmented GPT-2~\cite{radford2018improving} tokenizer as PyMT5, with about a vocabulary size of 50,000, and is the same architecture and hyperparameters as PyMT5 with 12 layers and 406M parameters. XPyMT5 was trained on 16 32GB Tesla V100 GPUs for 4 weeks, about 10 epochs total, using the same hyperparameters as reported by \citet{clement2020pymt5}. XPyMT5 was initialized with the English pre-trained BART~\cite{lewis2019bart} weights (with whitespace embeddings) and pre-trained using the BART de-noising objective for 5 weeks on the same hardware as above.

\subsection{Reformer/Performer}
We trained both Performer and Reformer models on the Python dataset described in \ref{sec:dataset} but without eWASH. Each training sample is a whole source code file literal normalization applied. We adapt the open-sourced model implementations,~\footnote{https://github.com/lucidrains/reformer-pytorch}\footnote{https://github.com/lucidrains/performer-pytorch} setting the architecture parameters of each to be as close to the same parameter count as XGPT-C as possible. Both used 12 layers, a context length of 4096, and 768 hidden dimensions. All other hyperparameters were unchanged from their default.

%using 12 layers of Transformer decoder with 768 hidden dimension. The context length is set to 4096. For locality-sensitive hashing, we set the bucket size as 64 and run with 4 hashes. The total number of model parameters is 116M, which is smaller than XGPT-C, but as close as we could get using the same projection layer for Q and K in multi-head attention.
% The pre-training period takes 2 weeks on sixteen 32GB Tesla V100 GPUs. 

%\subsection{Performer}
%We trained Performer model on the same dataset with Reformer. We adapt the open-sourced model implementation in Github,\footnote{https://github.com/lucidrains/performer-pytorch} using 12 layers of Transformer decoder with 768 hidden dimension. The context length is set to 4096, and we keep the original settings of their softmax approximations. The total number of model parameters is 125M. 

\section{Evaluation}

\begin{figure*}[htb]
    \includegraphics[width=0.95\textwidth]{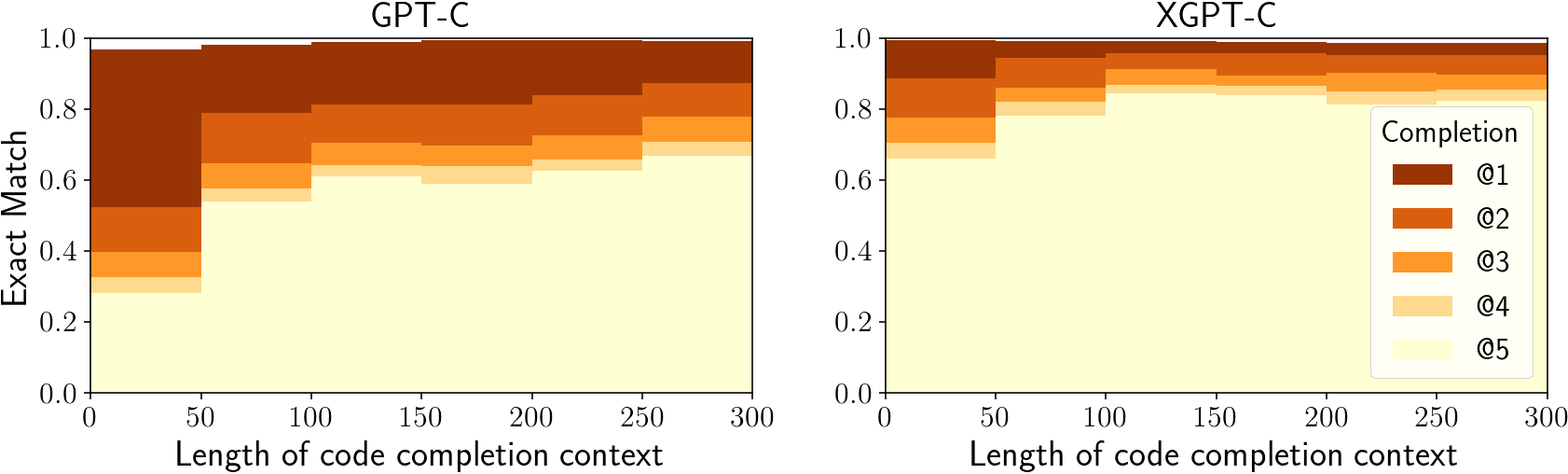}
    \caption{
    Comparing baseline GPT-C with XGPT-C in an offline evaluation of ExactMatch@1-5 code completion as a function of total token context length for the normalized literal scenario. Surprisingly, eWASH leads XGPT-C to benefit most over GPT-C at the shorter context lengths. XGPT-C also more exactly predicts tokens with longer context as well.
    }
    \label{fig:em-vs-context-length}
\end{figure*}

\begin{figure*}[htb]
    \includegraphics[width=0.95\textwidth]{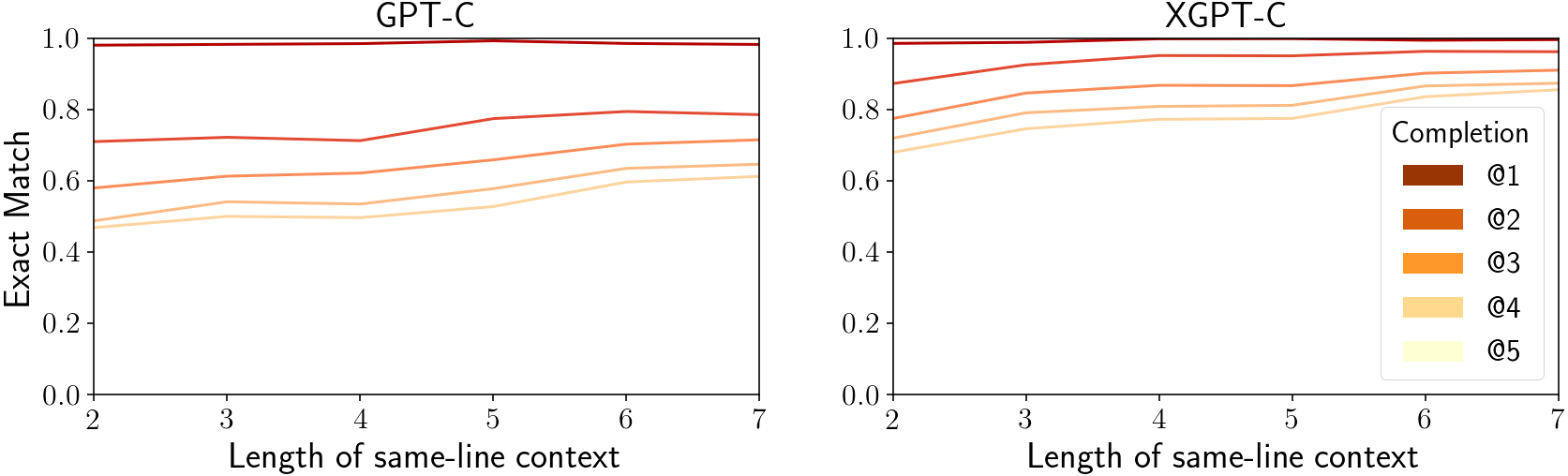}
    \caption{
    Comparing baseline GPT-C with XGPT-C in an offline evaluation of ExactMatch@1-5 code completion as a function of local, same-line context length for the normalized literal scenario. XGPT-C is better than GPT-C by all measures, showing how leveraging longer range context can most help developers even when a line being edited has a few tokens.
    }
    \label{fig:em-vs-sameline-context}
\end{figure*}

% Introduce evaluation tasks and motivate the evaluation metrics
% \XIAOYU{Reporting exact match and code completion configuration. Please feel free to remove if they are irrelevant or we are out of space}
\subsection{Metrics}
\label{sec:metrics}
The metrics we used to evaluate eWASH and thus XGPT-C and XPyMT5 are consistent with GPT-C and PyMT5 and other works in the literature. We report the longest common subsequence ROUGE-L as we expect in a developer tool scenario that users will want predicted code with the fewest edits. To that end, we also report the edit distance between the truth and hypothesis. In order to compare to other code completion models we report ExactMatch@N (EM) metrics~\cite{exactmatch}, which counts the fraction of exactly correct predictions of some length (@N). For method and docstring completion we report BLEU-4 and ROUGE-L metrics, but not exact matches as it is too strict to meaningfully interpret for longer source-target pairs. For method completion we also report the fraction of syntactically correct methods as judged by Python 3.8 syntax.

\subsection{Experimental Conditions}
\label{sec:experimental-conditions}
We aim to evaluate how well XGPT-C model can infer developers' true intents. We randomly selected 833 unique Python functions from the code completion test benchmark in CodeXGLUE~\cite{codexglue}, and, except for the first two tokens in each line, prompted the model at all other points inside the methods. The predictions are compared to the true continuation of the code. For method and docstring completion, the CodeSearchNet repositories and specific commit hashes were re-downloaded in order to extract the eWASH features in addition to the individual methods released. We will release this expanded CSN dataset and task to CodeXGLUE to improve its user-experience motivated metrics. Inference in all cases was performed with beam search with a beam width of 5.

%To perform a thorough evaluation, completion points are created from these function bodies in the following way: except for the first two tokens in each line of code, we create one completion point for each token together with its proceeding tokens in the same line. We conduct experiments on dataset as follows: In each experiment, we solve only one code completion problem. Specifically each code completion problem corresponds to exactly one completion point. By feeding the extend context and code context precede to the completion point to the model, the goal is to generate the Top-1 best suggestion. 

%Previous study~\cite{gptc} shows that code completion response time under 100 ms is necessary to avoid any feeling of delay or lag. To achieve this in our experiments, we adopt suggestion caching: at each completion point, suggestions are cached and queried when developer types a non-alphanumeric character.

\begin{table}[htb]
\small\centering
\resizebox{\columnwidth}{!}{
\begin{tabular}{rllllll} \toprule
& \multicolumn{5}{c}{\textbf{Exact Match}} \\  
& @1 & @2 & @3 & @4 & @5 & Total\\
\midrule
GPT-C top-1 & 96.0 & 68.5 & 56.3 & 49.6 & 46.3 & 63.1\\
 top-5 & 98.8 & 81.0 & 70.5 & 63.5 & 59.6 & 74.5\\
XGPT-C top-1 & 98.0 & 81.7 & 71.9 & 66.6 & 62.4 & 75.9\\
 top-5 & 98.9 & 94.3 & 87.7 & 83.1 & 79.7 & 88.7\\
\bottomrule
\end{tabular}
}
\caption{
Code completion evaluated on the CodeXGLUE test set by ExactMatch@1-5 and overall EM results for XGPT-C and GPT-C.
}
\label{tab:topk}
\end{table}

\subsubsection{Code Completion Evaluation Results}
% Results
As shown in Table~\ref{tab:eval_codecompletion}, eWASH allows XGPT-C to beat both the GPT-C baseline and the memory efficient transformers on all the metrics computed. About 10\% of our Python test files were greater than 1024 tokens in length, and evaluating separately on that subset yielded slight improvements of Performer/Reformer, but Performer only beat XGPT-C in terms of EM@5 at 55.9\%. These evaluations were performed for source code inside methods, as the eWASH technique follows the syntactic hierarchy used by developers. Note that the bottom lines are trained and evaluated on the normalized literal dataset. XGPT-C sees a large absolute increase in ExactMatch@5 of 13\% with normalized literals showing that, in addition to protecting user data, normalizing literals is an important part of a good IDE programming assistant.

\citet{hellendoorn2019code} showed that artificial evaluation scenarios are often much more forgiving than real-world scenarios. To better evaluate whether these models can predict a developer's intent we compute the ExactMatch@1-5 benchmark, described in Sec.~\ref{sec:metrics}, broken down by total token context and length of same-line context. Figure~\ref{fig:em-vs-context-length} shows EM@1-5 metrics for the normalized literal scenario binned by the context length for the completion for GPT-C (left) and XGPT-C (right). It is clear that in all measured cases eWASH allows XGPT-C to better predict exact matches. Perhaps most strikingly, the largest relative increase in EM occurs for shorter context lengths, so that the syntactic hierarchy hypothesis underlying eWASH appears most beneficial for context lengths well within the context window. 

\begin{table*}[htb]
\centering
\begin{tabular}{lllllllllll} \toprule
	\textbf{Model} & RL Prec. & Recall & F1 & \textbf{BLEU-4} & \textbf{Syntax (\%)} & \textbf{Model size} \\\midrule
%			\multicolumn{3}{c}{\textbf{ROUGE-L}} & \textbf{BLEU-4} & \textbf{Syntax (\%)} & \textbf{Model size} \\ \cmidrule{2-4}
%	                & Precision &  Recall & F1 \\ \midrule
PyMT5 Baseline & 0.33 & 0.46 & 0.35  & 0.27 & \bf 89\% & 406M\\	%final   
%Reformer & & & & & \\
%Performer & & & & & \\
XPyMT5 & \bf 0.52 & \bf 0.64 & \bf 0.55 & \bf 0.31 & 88\% & 406M &  \\
\bottomrule
\end{tabular}
\caption{Evaluation results for XPyMT5 multi-mode encoder-decoder model trained with extended hierarchical context and various baselines on the task of method generation given a natural language description. Model performance metrics are reported on CodeSearchNet test sample in Python programming language.} 
\label{tab:xpymt5-method-gen}	
\end{table*}

\begin{table*}[htb]
\centering
\begin{tabular}{llllllllll} \toprule
	\textbf{Model} & RL Prec. & Recall & F1 & \textbf{BLEU-4} & \textbf{Model size} \\\midrule
%			\multicolumn{3}{c}{\textbf{ROUGE-L}} & \textbf{BLEU-4} & \textbf{Model size} \\ \cmidrule{2-4}
%	                & Precision &  Recall & F1 \\ \midrule
CoTexT (CodeXGLUE leader) & & & & 0.197 & \\
PyMTBaseline & 0.32 & 0.37 & 0.32  & 0.28 &  406M\\	%final   
%Reformer & & & & & \\
%Performer & & & & & \\
XPyMT5 & \bf 0.58 & \bf 0.66 & \bf 0.66 & \bf 0.47 &  406M &  \\
\bottomrule
\end{tabular}
\caption{Detailed evaluation results for XPyMT5 model trained with extended hierarchical context and various baseline on code summarization task. Model performance metrics are reported on CodeSearchNet test sample in Python programming language.} 
\label{tab:xpymt5-docstring-gen}	
\end{table*}

Figure~\ref{fig:em-vs-sameline-context} shows the same EM metrics broken down by same-line context length, to test how much the most proximal tokens matter for prediction. We see the same overall benefit of eWASH in XGPT-C, and only a slow increase as a function of same-line context. The average line length in our data is 18 tokens, so with 7 tokens of same-line context, XGPT-C can complete 5 tokens exactly more than 80\% of the time while GPT-C can do so just shy of 60\% of the time. Again, this is very interesting as eWASH confers great benefit even when context lengths do not exceed the context window, and supports our hypothesis that user-defined syntax hierarchies are very important signals for predicting method bodies.

Modern IDE environments like Visual Studio Intellicode~\cite{intellicode} can present multiple predictions, which \citet{hellendoorn2019code} showed can improve real-world user acceptance. Table~\ref{tab:topk} shows the overall ExactMatch@1-5 metrics for code completion regardless of context length. XGPT-C is the clear winner again for all the EM metrics, boosting total exact matches by over 12\% for top-1 predictions and reaching 88.7\% overall for top-5 predictions. We interpret this to mean that eWASH will enable superior on-line user acceptance of code completions.

\subsection{Method Completion Evaluation Results}

We evaluate eWASH for method generation, illustrated in the top of Fig.~\ref{fig:ec-example}. Table~\ref{tab:xpymt5-method-gen} shows the comparison between XPyMT5 and PyMT5 and, and PyMT5 is superior in all the source-target comparison metrics. Syntax correctness is slightly lower, but the difference is not necessarily meaningful. The ROUGE-L metrics are dramatically improved, and is not necessarily surprising as XPyMT5 is conditioned on much more information than PyMT5. The syntax correctness of our fine-tuned models is slightly lower than the 92.1\% reported by \citet{clement2020pymt5}.

\subsection{Docstring Completion Evaluation Results}

Table~\ref{tab:xpymt5-docstring-gen} compares XPyMT5 to PyMT5 for docstring completion (or code summarization as CodeSearchNet removes the variable annotations). Again there is a large improvement in performance across all metrics, with a striking doubling of the ROUGE-L F1 score with eWASH features.

\section{Conclusions}

Inspired by the performance of transformer models, their limited context window size, and the especially long-range nature of source code as documents, we developed Extended Window Access by Syntax Hierarchy. Our hypothesis was that the syntax hierarchy imposed by developers is a real signal of importance in a task context, and that methods, containing most lines of code, are most dependent on the higher-level scopes of their file-level attributes. Our XGPT-C results for code completion supported this hypothesis, and, strikingly, offered most relative benefit for shorter context lengths. We showed with strict exact match metrics that eWASH allows a large relative improvement in code completion predictions. Finally we show dramatic improvement in method completion and code summarization with XPyMT5. eWASH can be applied to any programming language and in principal any language with hierarchical syntactic or stylistic structure. For this reason we believe eWASH to be a general purpose modeling approach for more optimally using finite context windows on structured documents, and could improve natural language understanding tasks as well. Further, any model, even the largest GPT-3 language model~\cite{brown2020language} can leverage the eWASH feature. Accompanying this manuscript we submit 3 new tasks to CodeXGLUE to bolster its user-experience motivated metrics: literal-normalized code completion, method-level code completion, and method/docstring completion conditioned on whole-file context.

%\Alexey{ can we hint/add a statement about possible applications to natural languages. For instance: Although this work focused on programming languages, improving natural language understanding and generation tasks with eWASH also seems possible. However, this would require large corpora of parsed text.}

%\section*{Acknowledgements}

\bibliography{references,emnlp2020}
\bibliographystyle{acl_natbib}

%\appendix
%\section{Example Appendix}
%\label{sec:appendix}
\newcommand{\appendixdefguard}{}
% uncomment to include appendix in the main document
\ifdefined\appendixdefguard

\else

\documentclass[11pt,a4paper]{article}
\usepackage[review]{emnlp2021}
\usepackage{times}
\usepackage{latexsym}
\renewcommand{\UrlFont}{\ttfamily\small}

% This is not strictly necessary, and may be commented out,
% but it will improve the layout of the manuscript,
% and will typically save some space.
\usepackage{microtype}
\usepackage{makecell}
\usepackage{booktabs}
\usepackage{geometry}
\usepackage{graphicx}
\usepackage{caption}
\usepackage{xcolor}
\usepackage{minted}
\usepackage{array}
\usepackage{pifont}

\usepackage{tikz}
\usetikzlibrary{positioning}
\usepackage{subfig}

% Author information can be set in various styles:
% For several authors from the same institution:
% \author{Author 1 \and ... \and Author n \\
%         Address line \\ ... \\ Address line}
% if the names do not fit well on one line use
%         Author 1 \\ {\bf Author 2} \\ ... \\ {\bf Author n} \\
% For authors from different institutions:
% \author{Author 1 \\ Address line \\  ... \\ Address line
%         \And  ... \And
%         Author n \\ Address line \\ ... \\ Address line}
% To start a seperate ``row'' of authors use \AND, as in
% \author{Author 1 \\ Address line \\  ... \\ Address line
%         \AND
%         Author 2 \\ Address line \\ ... \\ Address line \And
%         Author 3 \\ Address line \\ ... \\ Address line}

\title{Long-Range Modeling of Source Code Files with eWASH: Extended Window Access by Syntax Hierarchy}

\author{First Author \\
  Affiliation / Address line 1 \\
  Affiliation / Address line 2 \\
  Affiliation / Address line 3 \\
  \texttt{email@domain} \\\And
  Second Author \\
  Affiliation / Address line 1 \\
  Affiliation / Address line 2 \\
  Affiliation / Address line 3 \\
  \texttt{email@domain} \\}

\begin{document}

\definecolor{mygray}{RGB}{230, 230, 230}
\newcommand*\rot{\rotatebox{90}}
\newcommand*\OK{\ding{51}}

\newcommand{\python}{\texttt{python}}
\newcommand{\ourmodel}{\textsc{PyMT5}}

\fi

%\aclfinalcopy % Uncomment this line for the final submission
%\def\aclpaperid{***} %  Enter the acl Paper ID here

%\setlength\titlebox{9cm}
% You can expand the titlebox if you need extra space
% to show all the authors. Please do not make the titlebox
% smaller than 5cm (the original size); we will check this
% in the camera-ready version and ask you to change it back.

%\newcommand\BibTeX{B\textsc{ib}\TeX}

\maketitle

\appendix

\section{Appendix}
\label{sec:appendix}

\subsection{Code Completion Above and Below the Context Window}

The low performance of the memory efficient transformers in the main text is puzzling, so we decomposed our test set on code completion into the 90\% with fewer than 1024 tokens (the window of GPT-C and XGPT-C), and 10\% with more than 1024 tokens.

\begin{table*}[htb]
\centering
\begin{tabular}{lllll} \toprule
	\textbf{Model}  & \bf ROUGE-L Prec. & \bf Recall & \textbf{Edit dist.} & \textbf{EM@5 (\%)}\\\midrule
%			\multicolumn{2}{c}{\textbf{ROUGE-L}} & \textbf{Edit dist.} & \textbf{EM@5 (\%)}  & \textbf{Size} \\ \cmidrule{3-4} \cmidrule{8-10}
%	                & & Precision &  Recall \\ \midrule
Performer & 0.73 & 0.82 & 89.6 & 47.9 \\
Reformer & 0.76 & 0.84 & \bf 93.6 & \bf 55.9 \\
XGPT-C & \bf 0.85 & \bf 0.94 & 90.9 & 49.3\\\midrule
GPT-C, Norm Literals & 0.90 & 0.94 & 90.9 & 49.3 \\	%final  
XGPT-C, Norm Literals  & \bf 0.91 & \bf 0.96 & \bf 94.1 & \bf 63.9\\	
\bottomrule
\end{tabular}
\caption{Evaluation results comparing XGPT-C on code completion from sampled methods from files with more than 1024 tokens. Model performance metrics are reported on test samples from the CodeXGLUE code completion task as described in the main text. 
}
\label{tab:more-tokens}	
\end{table*}

\begin{table*}[htb]
\centering
\begin{tabular}{lllll} \toprule
	\textbf{Model}  & \bf ROUGE-L Prec. & \bf Recall & \textbf{Edit dist.} & \textbf{EM@5 (\%)}\\\midrule
%			\multicolumn{2}{c}{\textbf{ROUGE-L}} & \textbf{Edit dist.} & \textbf{EM@5 (\%)}  & \textbf{Size} \\ \cmidrule{3-4} \cmidrule{8-10}
%	                & & Precision &  Recall \\ \midrule
Performer & 0.68    &    0.80  &    84.9    &   40.5 \\
Reformer & 0.70     &    0.81  &     86.0  &    45.1\\
XGPT-C & \bf 0.85  &  \bf 0.93  &   \bf 90.8  &   \bf 49.4 \\\midrule
GPT-C, Norm Literals &  0.81    &      0.89  &  91.1  &  45.3    \\	%final  
XGPT-C, Norm Literals  & \bf 0.90 & \bf 0.97  & \bf 93.6  &  \bf 62.2 \\	
\bottomrule
\end{tabular}
\caption{Evaluation results comparing XGPT-C on code completion from sampled methods from files with more than 1024 tokens. Model performance metrics are reported on test samples from the CodeXGLUE code completion task as described in the main text. 
}
\label{tab:fewer-tokens}	
\end{table*}

\ifdefined\appendixdefguard

\else
\end{document}
\fi

\end{document}